# A STUDY ON NON-DESTRUCTIVE METHOD FOR DETECTING TOXIN IN PEPPER USING NEURAL NETWORKS

M.Rajalakshmi[1] and Dr.P. Subashini[2]

Research Scholar, Department of Computer Science, Avinashilingam Deemed University for Women, Coimbatore.
rajisaravanan25@gmail.com

Associate professor, Department of Computer Science, Avinashilingam Deemed University for Women, Coimbatore.
mail.p.subashini@gmail.com

## ABSTRACT

*Mycotoxin contamination in certain agricultural systems have been a serious concern for human and animal health. Mycotoxins are toxic substances produced mostly as secondary metabolites by fungi that grow on seeds and feed in the field, or in storage. The food-borne Mycotoxins likely to be of greatest significance for human health in tropical developing countries are Aflatoxins and Fumonisins.*

*Chili pepper is also prone to Aflatoxin contamination during harvesting, production and storage periods. Various methods used for detection of Mycotoxins give accurate results, but they are slow, expensive and destructive. Destructive method is testing a material that degrades the sample under investigation. Whereas, non-destructive testing will, after testing, allow the part to be used for its intended purpose.*

*Ultrasonic methods, Multispectral image processing methods, Terahertz methods, X-ray and Thermography have been very popular in nondestructive testing and characterization of materials and health monitoring. Image processing methods are used to improve the visual quality of the pictures and to extract useful information from them. In this proposed work, the chili pepper samples will be collected, and the X-ray, multispectral images of the samples will be processed using image processing methods.*

*The term "Computational Intelligence" referred as simulation of human intelligence on computers. It is also called as "Artificial Intelligence" (AI) approach. The techniques used in AI approach are Neural network, Fuzzy logic and evolutionary computation. Finally, the computational intelligence method will be used in addition to image processing to provide best, high performance and accurate results for detecting the Mycotoxin level in the samples collected. This research paper gives an overview of the ongoing research in non-destructive methods for finding toxins in chili pepper by making a comparative study of the previous works.*

## 1.INTRODUCTION

Accessing high quality and safe food is one of the most important things to the public anxiety in recent decades. Food safety implies that the food includes safe levels of the substances, which does not include toxins and contaminants cause injurious to human health. Food quality includes characteristics such as nutritional value and texture for the food to be preferred by the consumers. Many countries impose strict control on the standards of food contents which directly affect the health and quality. It is essential to all the countries to assure the safety and quality of the imported food to protect their domestic consumers. In addition to the domestic consumers more attention should be given to the safety and quality of export food in the global market. Diseases arising from the food lead threat to human health and bring decrease in the economic productivity of respective countries.

### 1.1. Non-destructive methods

Non-destructive methods are a part of quality control function and complementary to other established methods. Non-destructive testing is the testing of materials, for surface or internal flaws or metallurgical condition, without interfering in any way with the reliability of the material or its suitability for service. This method is suitable for testing both sampling units for the individuals and entire production units for the quality control centre. Development of high technology and evolution of equipments brings huge quality control applications in the lead-time of production sectors. To facilitate production the application of this method requires certain degree of skills to obtain more information about the manufactured product and to improvise the product requires consequent feedback from the quality control centre. Non-destructive methods are not only used for rejecting substandard materials and also an assurance given to the materials that are evidently good. There is no single method around which a black box may be built to satisfy all requirements in all circumstances. After the detailed study of the previous research works on non-destructive methods, it is proposed to follow the various steps referred in the following Fig: 1[28].

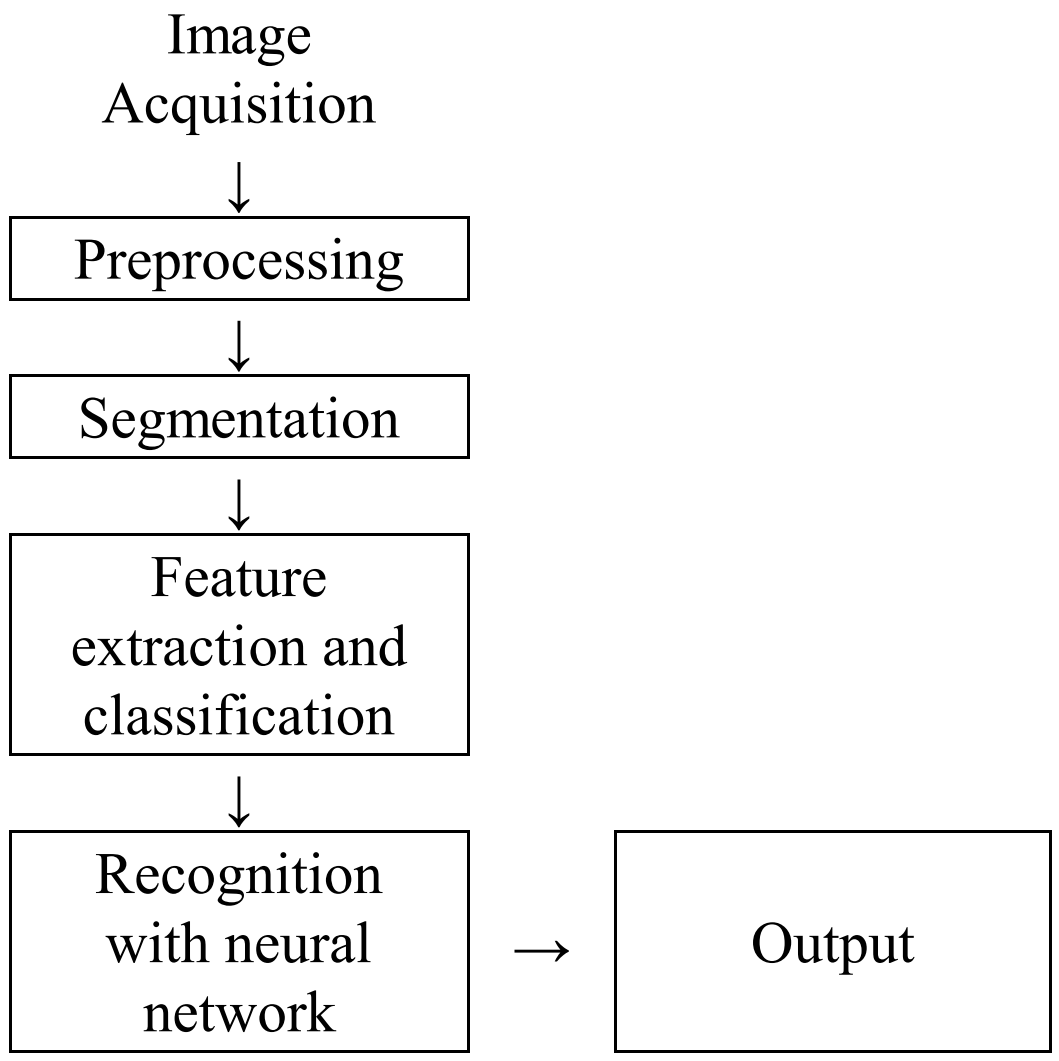

Fig. 1. Proposed system.

## 2.METHODOLOGY

### 2.1.Preprocessing

The key function of preprocessing is to improve the image such that it increases the chances for success of other processes. The preprocessing techniques are used for enhancing the contrast of the image, removal of noise and isolating the objects of interest in the image.

Morten Amgren[25], proposed a work on classification of grain samples, under the assumption of isotropic noise, the data can be reduced to k-1 dimensions without loss of classification performance by projecting onto the hyper plane spanned by the class means. Data is hence reduced to two dimensions.

In 1985, B.L.Upchurch, E.S. Furgason, G.E.Miles and F.D.Hess[4], developed a work on ultrasonic measurements for detecting damage on agricultural products, to estimate the power spectrum they used blackman-tukey method. To prevent leakage errors, a Hanning window of length 70 data points is used to window the data.

In 1999, Zeev Schmilovitch, Aharon Hoffman, Haim Egozi, Rachel Ben-zvi, Zvi Bernstein and Victor Alchanatis

[24],determined maturity of fresh dates by near infrared spectrometry. For that they used first derivative as preprocessing method. The desired NIR model should have a low error of prediction, with as few factors as possible, and observation of this criterion led to a calibration procedure which used a model based on the first derivative of the spectra and four-factor PLS regression. The PLS regression models relate the first derivative NIR spectra to the water and TSS contents in each tested date.

In 2005, Marina Cocchi, Maria Corbellini, Giorgia Foca, Mara Lucisano, M Ambrogina Pagani, Lorenzo Tassi and Alessandro Ulrici, [17], developed a work on classification of bread wheat flours. The wheat spectra were pretreated using detrending, standard normal variate, first and second derivatives.

In 2007, Yong He, Xiaoli Li and Xunfei Deng [23], a new approach for discrimination of varieties of Tea was developed. To reduce the noise, such as baseline excursion, high frequency noise and dispersion, the smoothing way of Savitzky was used. Another type of preprocessing is multiplicative scatter correction (MSC) which is used to modify the additive and multiplicative effects in the spectra. Due to longer light traveling path, more light is absorbed which causes a parallel translation of the spectra. This kind of deviation could be eliminated by MSC-technique.

In 2008, J.M.Hernandez-Hierro, R.J.Garcıa-Villanova and I.Gonzalez-Martına[14], in their work on analysis of mycotoxins contaminated red paprika, the effects of scattering were removed using Multiplicative Scatter Correction(MSC) , Standard Normal Variate(SNV) , DeTrend (DT), or SNV-DT.

In 2009, Pelin Beriat[26], developed a Machine vision for textured foods. To remove impulsive noise, chili pepper images are filtered with median filter with size of 7x7.

In 2009, SmitaTripathi and ,H.N.Mishra[21], discussed about estimation of aflatoxin B1 in red chili powder, they used Minimum maximum normalization, Constant offset elimination and straight line subtraction based upon the lowest values of Root Mean Square Error of Cross Validation(RMSECV).

In 2009, V.Fernandez-Lbanez, A. Soldado, A. Martinez-Fernandez and B.De La roza-Delgado [22], detected aflatoxin $B_1$ in maize and barley. Two different strategies were tested as spectral data pretreatment, the first was a null option, where log (1/*R*) was not scatter corrected and the second was a standard normal variate and detrending (SNVD) pretreatment for scatter correction. In addition to that, three derivative mathematical treatments were tested to develop NIR calibrations. The best discriminant model for each grain type was selected using the lowest standard error of cross-validation (SECV) and highest coefficient of determination in cross-validation ($r^{2)}$.

In 2010, Atas Moses, Jasmine Help and Alptekin Temizel[3], proposed a work on classification of hyperspectral images of chili peppers with aflatoxins , 3x3 median filter was used to remove noise from chili pepper images.

In 2010, Geraudie V., Roger J.M., Ferrandis J.L., Gialis J.M., Barbe P., Bellon Maurel V., and Pellenc .R [10], proposed a revolutionary device for predicting grape maturity, they used averaging, Normalization and transformation for removing immature grape images.

In 2010, J.Liu, X.Y.Li, P.W.Li, W.Wang, J.Zhang, Z.Zhang and P.Liu [13], presented a work on nondestructive detection of moldy chestnut. Preprocessing of raw spectra is necessary in NIR spectroscopy modeling, in order to eliminate the interference caused by random noise, baseline drift, temperature and other factors. The Vector normalization (VN), first derivative (FD), first derivative and vector normalization (FD and VN), second derivative (SD), second derivative and vector normalization (SD and VN) plus no preprocessing (NP) were compared, and found that FD and VN method was more effective and accurate.

## 2.2 Segmentation

Segmentation is a process in which the given input image is partitioned into its constituent parts or objects. The key role of segmentation in the mechanical component classification is to extract the boundary of the object from the background. The output of the segmentation stage usually consists of either boundary of the region or all the parts in the region itself.

Morten Amgren[25], proposed a work on classification of grain samples, a scalar threshold applied to the first principal component image to obtain segmented image.

In 2002, N.Aleixos, J.Blasco, F.Navarron and E.Molto[19], in their work on multispectral inspection of citrus, each image is binarised with a threshold and the boundaries of each of the objects are extracted, applying a convolution mask. Thresholding the infrared image provided a simple and effective means for segmenting the fruit from the background.To obtain size and shape estimation, the following parameters are calculated. They are centroid, maximum and minimum diameter, perimeter and circularity.

In 2008, J.M.Hernandez-Hierro, R.J.Garcıa-Villanova and I.Gonzalez-Martına[14], discussed about the analysis of mycotoxins contamination in red paprika. Aflatoxins quantization was accomplished by modified partial least squares (MPLS) regression method to obtain the NIR equations for all the parameters studied. In MPLS, the NIR residuals at each wavelength, obtained after each factor has been calculated are standardized before calculating the next factor. When developing MPLS equations, cross-validation is recommended in order to select the optimal Modified Partial Least Squares (MPLS) algorithm with the help of regression method.

In 2009, Di Cui, Qin Zhang, Minzan Li, Youfu, Zhao and Glen L.Hartman [9], in their work on detection of soybean rust using a multispectral image sensor, they utilized threshold function to separate the soybean severity based upon infected area separation, lesion color identification and rust severity quantification. Laboratory- scale test results verified that among the set of defined parameters the RSI (Rust Severity Index) positively correlate with the severity of rust infection.

In 2009, SmitaTripathi and ,H.N.Mishra[21], discussed about estimation of aflatoxin B1 in red chili powder, they used two important methods generally employed in chemometrics for multi component analysis are Least Squares Support Vector Machines(LS-SVM) and Partial Least Squares(PLS), however PLS regression method is most widely used in many cases involving linear relationship between the analytical signal and the property to be determined.

In 2010, Alberto Guillen, F.G.Del Moral, L.J.Herrera, G.Rubio, I.Rojas, O.Valenzuela and H.Pomares [1], developed a new approach on classification of pork with neural networks. They analyzed data using mutual information theory. Estimation of the joint probability density functions between X and Y is done by using k-nearest neighbours concept which is needed to estimate mutual information between two group of variables. The determination of which reflectance values are significant to determine the breed of pork was calculated by estimating the mutual information between every input variable and the breed.

In 2010, Ana Herrero- Langreo, Loredana Lunadei, Lourdes Lleo, Belen Diezma and Margarita Ruiz-Altisent [2], in their work on monitoring peach fruit ripeness, they used triangle algorithm for segmentation. When compared to the Otsu method the performance improve from 90% to 96% for the triangle algorithm segmented images.

In 2010, Devrim Unay, Bernard Gosselin, Olivier Kleynen, Vincent Leemans, ,Marie-France Destain and ,Olivier Debeir[8], introduced, a new automatic fruit grading system by multispectral machine vision. Automatic segmentation of defects is achieved by sequential application of two previously introduced solutions. First, candidate defect regions are segmented through pixel-wise classification by a MLP-based method. Then, these segmentations are refined by excluding stem and calyx regions–natural parts of fruit that show similar spectral characteristics with some defects–found by a support vector machine (SVM) based method.

## 2.3. Feature Extraction and Classification

Feature extraction is a process to select important characteristics of an image or object. Image classification maps different regions or segments into one of several objects, each identified by a label.

Morten Amgren[25], proposed a classification of grain samples, simple K-means algorithm is applied to the training set conducting unsupervised classification. Classification is further used as labels to train the supervised model by using LDA and QDA.

In 1985, B.L.Upchurch, E.S. Furgason, G.E.Miles and F.D.Hess[4], developed a work on ultrasonic measurements for detecting damage on agricultural products, to distinguish between undamaged and damaged apple tissue depends on a redistribution of the water within the tissue. Scanning Electron Micrographs (SEM) showing the differences within the intercellular spaces between undamaged and damaged apple tissue are presented. The change in tissue was detected by ultrasonic transducers.

In 1998, Knut Kvaal, Jens Petter Wold, Ulf G. Indahl, Pernille Baardseth and Tormod Næs[15], in their work on multivariate feature extraction from textural images of bread, they compared five different methods of extracting features from textural images in food by multivariate modeling of the sensory porosity of wheat baguettes. The methods treated are the angle measure technique (AMT), the singular value decomposition (SVD), the auto correlation and auto covariance functions (ACF) and the so-called size and distance distribution (SDD) method. The methods will be tested on equal basis and the modeling of sensory porosity from extracted features is done using principal component regression (PCR) and partial least square re gression(PLS). The results show that all the methods are suited to extract sensory porosity but the AMT method proves to be the best in this case.

In 1999, Zeev Schmilovitch, Aharon Hoffman, Haim Egozi, Rachel Ben-zvi, Zvi Bernstein and Victor Alchanatis [24], developed a semi-automatic system for maturity determination of fresh dates. Processed spectra such as the first derivative, log (1/R), and its first and second derivatives were analyzed by principal component analysis, partial least squares (PLS) and Multi-linear regressions by using the Spectra Metrix and Light Cal software packages. Dates with water content < 66% or with sugar content > 30.5% are regarded as mature and will ripen properly. The prediction results had a Standard Error of Prediction (SEP) of 1% for both TSS and water contents. The correlation coefficient, R between the predictions of TSS and water contents based on NIR models and on routine destructive laboratory tests was 0.9.In 2000, C.S.Burks, F.E. Dowell and F.Xie [6], they measured fiq quality by using near-infrared spectroscopy. When analyzing data, a cutoff was selected that resulted in the greatest number of correctly classified figs. spectra were analyzed using Partial Least Squares (PLS) and GRAMS software.

In 2002, N.Aleixos, J.Blasco, F.Navarron and E.Molto[19], in their work on multispectral inspection of citrus, to detect defect Bayesian discriminant model is then generated and stored in a look up table. Inspection of the image of the strips is partially based on raster-scan algorithm.

In 2004, C.Karunakaran, D.S.Jayas and N.D.G.White[5], identified the damages of wheat kernels using X-ray images, they carried out DISCRIM procedure of the statistical classifier. Four layer back propagation neural network was used to identify uninfected and infected wheat kernels. There was no significant difference between the statistical and neural network classifiers for the identification of uninfected and infected wheat kernels.

In 2005, Chun-Chieh Yang, Kuanglin Chao, Yud-Ren Chen and howard L.Early[7], developed a work on identification of diseased chicken using multispectral images. They used Principal Component Analysis (PCA) algorithm. In this work relative reflectance intensity at individual wavelengths, ratio of intensities between pairs of wavelengths and intensity combinations based on principal component analysis were analyzed. Classification And Regression Tree (CART) decision tree algorithm was used to determine the threshold value and found satisfiable.

In 2005, Marina Cocchi, Maria Corbellini, Giorgia Foca, Mara Lucisano, M Ambrogina Pagani, Lorenzo Tassi and Alessandro Ulrici, [17], discovered a new approach for classification of bread wheat flours. They employed SIMCA and Wavelet packet transform for efficient pattern recognition (WPTER) algorithms. The wheat samples are categorized into four categories. They are improver wheat, superior bread making wheat, ordinary bread making wheat, wheat for biscuits. Due to the high overlap of the two intermediate quality classes, it was not possible to classify all the data set signals. However, when considering only the two extreme categories, an acceptable degree of class separation can be gained after feature selection by WPTER. Moreover, this approach allowed locating the NIR spectral regions that are mainly involved in the assignment of the wheat flour samples to these two quality categories.

In 2007, Ran Zhou and Yunfei Li [20], in their work on texture analysis of Magnetic resonance image for predicting the firmness of huanghua pears, A feed forward neural networks with back-propagation algorithm was designed. Log-sigmoid transfer function was used with eight inputs, one output. Performance measured by Mean Absolute Error (MAE).Numerical analysis was used to extract information on the changes in macroscopic structure from the images. Firmness of pears was determined through Pearson's correlation analysis.

In 2007, Yong He, Xiaoli Li and Xunfei Deng [23], in their research they made quantitative analysis for the varieties of tea, by combining the WT, PCA and BP-ANN techniques. A relation was established between reflectance spectra and varieties of tea. The BP-ANN shows an excellent data prediction performance especially after the spectral data were pre-treated by WT and PCA. And the recognition rate of 100% was achieved. With the applying of the principal components analysis (PCA), the building and training of ANN model becomes feasible.

In 2008, J.M.Hernandez-Hierro, R.J.Garcıa-Villanova and I.Gonzalez-Martına[14], described a complete system for analysis of mycotoxins contamination in red paprika. Spectra of pepper were treated using Principal Component Analysis (PCA) algorithm. The prediction capacity of the model was assessed using the ratio performance deviation (RPD) parameter. Data of p (level of significance), Residual mean, Root Mean Standard Error (RMSE) indicates that the procedure is suitable.

In 2008, LI Xiaoli,HE Yong, QIU Zhengjun, WU Di[16], proposed a work on texture discrimination of green tea categories based on least squares support vector machine, each image was generated based on texture analysis techniques including grey level co-occurrence method and texture filtering. Unsupervised cluster analysis was conducted using principal component analysis. Least squares support vector machine classifier was developed based on the optimized texture features. Finally concluded that texture discrimination of green tea categories based on multi-spectral image technology is feasible.

In 2009, Pelin Beriat[26], developed a Machine vision for textured foods. He was used two approaches for extracting the most discriminative features are Statistical approach and LDB approach. In statistical approach, a combination of First Order Statistical (FOS) features and Gray Level Cooccurrence Matrix (GLCM) features are extracted. The second approach for feature extraction is 2D Local Discriminant Bases (LDB) approach. A standard LDA classifier is used for evaluating the classification accuracy of the selected features.

In 2009, SmitaTripathi and,H.N.Mishra[21], discussed about estimation of aflatoxin B1 in red chili powder. They used two important methods generally employed in chemo metrics for multi component analyses are Least Squares Support Vector Machines (LS-SVM) and Partial Least Squares (PLS). PLS regression method is most widely used in many cases involving linear relationship between the analytical signal and the property to be determined. The overall results demonstrated that FT-NIR spectroscopy can be used for rapid, nondestructive quantification of aflatoxin B1 in red chili powder.

In 2009, V.Fernandez-Lbanez, A. Soldado, A. Martinez-Fernandez and B.De La roza-Delgado [22], developed a application to detect aflatoxin $B_1$ in maize and barley. They used discriminant analysis with the Partial Least Squares (PLS) in addition to the multivariate statistical method. In this work three chemo metrics models were evaluated. They are discriminate maize contaminated and non-contaminated ABF1, discriminate barley contaminated and non-contaminated ABF1, discriminate cereal grain.ABF1 occurs in small concentrations, it is not likely that NIRS can detect ABF1 directly. However contamination by aflatoxins affects other chemical and optical properties of whole kernels that can be detected with Vis or NIR spectroscopy. The external validation samples were classified correctly.

In 2010, Alberto Guillen, F.G.Del Moral, L.J.Herrera, G.Rubio, I.Rojas, O.Valenzuela and H.Pomares [1], developed a work on classification of pork with neural networks, mutual information theory was used to deal the determination of which wavelengths of the spectrum are more relevant for the classification. The RBFNNs generated with the FPCFA algorithm obtain the best results in comparison with the FCM and the SVMs. The reason is the addition of the supervising parameter and the combination of possibilistic and fuzzy partitions. The RBFNNs get their maximum classification performance using a variable, and if the number of variable is increased, they perform well but not as much as with fewer variables. RBFNNs are local models; they perform better in small dimensions whereas SVMs have an exceptional performance.

In 2010, Ana Herrero- Langreo, Loredana Lunadei, Lourdes Lleo, Belen Diezma and Margarita Ruiz-Altisent [2], in their work on monitoring peach fruit ripeness, they used a non supervised classification according to Ward's method in order to define a ripeness reference class based on histograms of the ratio images of a learning data set. Ward's classification methods was applied by computing the matrix of Euclidean distances between each pair of individuals, grouping the closest individuals and hierarchically merging groups whose combination gave the least

Ward linkage distance.

In 2010, Atas Moses, Jasmine Help and Alptekin Temizel[3], proposed a work on classification of hyperspectral images of chili peppers with aflatoxins, they used Principal Component Analysis(PCA), and Self –Associative neural network as classifier.

In 2010, DevrimUnay, Bernard Gosselin, Olivier Kleynen, Vincent Leemans,Marie-France Destain and ,Olivier Debeir[8], introduced automatic grading of fruits by multispectral machine vision. They used Sequential Floating Forward Selection (SFFS) which is a heuristic and greedy algorithm that starts with an empty feature subset. In this work Statistical measures do not take relative relations of gray values into account whereas textural features consider pixels in pairs and thus capture spatial dependence of gray values that contribute to perception of texture. Geometric moments of Hu are textural features that are widely used in pattern recognition. Four statistical and one syntactical classifiers are used i.e., Linear Discriminant Classifier(LDA), K-Nearest Neighbor classifier(KNN), Fuzzy KNN and Support Vector Machines(SVM) with one against all strategies, and c4.5. Performances of the classifiers study are calculated by overall accuracy, producer's accuracy, user's accuracy and cohen's kappa statistic. Improved recognition rates can be achieved with cascaded classifiers.

In 2010, Geraudie V., Roger J.M., Ferrandis J.L., Gialis J.M., Barbe P., Bellon Maurel V., and Pellenc .R [10], proposed a revolutionary device for predicting grape maturity. The spectra were filtered by Principal Component Analysis (PCA) and by specific filters to remove outliers. To extract and to generate calibration models, Multiple Linear Regression (MLR) and leave one out cross validation were used. Standard Error of Cross Validation (SECV) and other measurements were calculated. The results show very satisfactory performances with regard to the grape composition and the ripeness monitoring.

In 2010, J.Liu, X.Y.Li, P.W.Li, W.Wang, J.Zhang, Z.Zhang and P.Liu [], proposed a work on nondestructive detection of moldy chestnut, they utilized seven classification models. They were established based on the data derived from different sample groups, using a supervised pattern recognition method. The methods were single linkage (SL), complete linkage (CL), average linkage (AL), weighted average Linkage (WAL), median algorithm (MA), centroid algorithm (CA) and Ward's algorithm (WA). A set of spectra for the three chestnut groups that is, sound chestnuts without mildew, slight mildew and severe mildew were selected as the training set and the distances between the spectra were calculated to function, as a standard for screening moldy Chestnuts without knowing the mildew conditions. Among the other Ward's algorithm performs well. Analyses were carried out by using the OPUS software.

In 2010, Musa Atas, alptekin Temizel and Yasemin Yardmci[13], developed a approach for classifying aflatoxin contaminated chili pepper. They used Principal Component Analysis (PCA), Self Associative Neural Network(Auto associative ANN) algorithm for feature selection. Spectral images based on the texture of the gray level co-occurrence matrix. For classification they used Feed-Forward Back Propagation Artificial Neural Network Classifier. For normalization Z-score method was used to obtain best results.

In 2011, Hongfei Lu, Hong Zheng, Ya Hu, Heqiang Lou and Xuechen Kong [12] developed a new method to sort red bayberries based on the presence of bruises. They used principal component- support vector machine and support vector machine models combined with fractal analysis and compared with classification models based on RGB intensity values. The performance of the SVM model in terms of iteration time and the number of support vectors was better than the PC-SVM model. Therefore the SVM model based on fractal analysis is recommended for detecting bruises on red bayberries.

In 2011, H.Kalkan, P.Beriat, Y.Yardimci and T.C.Pearson [11], detected contaminated hazelnuts and ground red chili pepper by using two dimensional Local Discriminant Bases (LDB) and Linear Discriminant Analysis (LDA) algorithm. LDB Algorithm is used to detect the location of the discriminative features in the multispectral data space. Feature extraction step includes two consecutive pruning operations first on the feature tree generated along the spectral and second on the tree generated on the spatial frequency axis. For feature selection three simple algorithms are used. They are Fisher based, Wrapper based and Forward Selection. Among them forward selection gives subset including the most discriminative feature. Then this subset is extended incrementally with new features. For classification Linear Discriminant Analysis (LDA) was used. Candidate features were either all fed directly into the classifier or reduced by Principal Components Analysis (PCA) and then fed into the classifier. Mean aflatoxin level decreased to 22.85 ppb from the group average of 38.26 ppb by separation of the aflatoxin contaminated pepper samples.

## 3.CONCLUSION

In this paper, the ongoing research work carried out in non-destructive methods to detect toxin in chili pepper. This paper tried to bring out the present status of non-destructive methods. Although each of the methods summarized above have their own superiorities and drawbacks, and their recognition results of different methods seem very successful. From all the above, this paper conclude that multispectral images of chili pepper may works well with less amount of preprocessing. Most of the recognition accuracy rates reported is over 78%. From the above discussion, the proposed system infers that by using filters for preprocessing and Principal Component Analysis (PCA) for Feature extraction of chili pepper gives more accurate results. The proposed system also infers that toxin detection using computational intelligence techniques shows a quite good experimental result. But all the improved algorithms were based on traditional training algorithm. For this reason, the proposed system will use Support Vector Machines algorithm (SVM) to give more accurate and better results.

## Table 1 .Comparative study on non-destructive methods

| Year | Author | Method applied | Preprocessing | Segmentation | Feature extraction and classification | Results and inferences |
|---|---|---|---|---|---|---|
| 1985 | B.L.Upchurch et al, | Ultrasonic | Hanning window of length 70 data points | - | Scanning electron micro graphs, ultrasonic | - |
| 1998 | Knut Kvaal et al, | Multivariate analysis | - | - | Principal component regression, Partial least square regression. | Correlation coefficient is 0.90 |
| 1999 | Zeev Schmilovitch et al, | Near Infrared spectrometry | First derivative | - | Principal component analysis, Partial least square, Multi linear regression | Correlation coefficient is 0.9 |
| 2000 | C.S.Burks et al, | Near Infrared spectroscopy | - | - | Partial least square, GRAMS software | Accuracy is 83 to 100 |
| 2002 | N.Aleixos et al, | Multi spectral | - | Thresholding | Bayesian discriminant model, Raster scan algorithm | Accuracy is 93 |
| 2004 | C.Karunakaran et al | X-Ray | - | - | DISCRIM procedure, Back propagation neural network | Accuracy is 73 to 86 |
| 2005 | Chun-Chieh Yang et al, | Multispectral | - | - | Principal component analysis, classification and regression tree algorithm | Accuracy is 95.7 to 100 |
| 2005 | Marina Cocchi et al, | Near Infrared spectra | De-Trending, Standard Normal Variate, first and second derivative | - | SIMCA algorithm, Wavelet Packet Transform for efficient pattern recognition | sensitivity and specificity values >= 80 |
| 2007 | Ran Zhou et al, | Magnetic resonance | - | - | Feed forward neural network with back propagation, log-sigmoid transfer function | Correlation coefficient is 0.969 |
| 2007 | Yong he et al, | Near Infrared spectroscopy | Multiplicative Scatter Correction | - | Principal component analysis, back propagation neural network | Accuracy is 100 |
| 2008 | J.M.Hernandez-Hierro et al, | Near Infrared spectroscopy | Multiplicative Scatter Correction, Standard Normal Variate, De-Trend or SNV-DT | Modified partial least squares algorithm | Principal component analysis | Multiple correlation coefficient is 0.955 |
| 2008 | LI Xiaoli et al, | Multispectral | - | - | Principal component analysis | Accuracy is 100 |
| 2009 | Di Cui et al, | Multispectral | - | Thresholding function, Rust severity index | - | rust severity index increases from 0.0 to 17.0 |
| 2009 | Pelin Beriat | Hyper Spectral Image | Median Filter with size of 7x7 | - | Statistical, linear discriminant bases, local discriminant bases | Accuracy is 79.17 |
| 2009 | Smitha Tripathi | FT-NIR | Minimum maximum normalization, Constant offset elimination and straight line subtraction | Least Squares Support Vector Machine, Partial least square | Least Squares Support Vector Machine | Accuracy is 100 |
| 2009 | V.Fernandez-Lbanez et al, | Near infrared spectroscopy | Standard Normal Variate and De-Trending, Null option,Three Derivative | - | Discriminant Analysis with partial least square | Accuracy is 100 |
| 2010 | Alberto Guillen et al, | Near infrared spectroscopy | - | Entropy based joint probability density function, K-nearest neighbor | Radial bases function neural networks and support vector machine, mutual information theory | Accuracy is 99 |
| 2010 | Ana Herrero – Langreo et al, | Multispectral | - | Triangle algorithm | Ward's method | Accuracy is 80 |
| 2010 | Atas Moses et al, | Hyper spectral | 3x3 median filter | - | Principal Component | Accuracy is |
| | | image | | | analysis, Self associative neural network | 85 |
| 2010 | Devrim Unay et al, | Multispectral | Band-pass filters | MLP Method | Sequential Floating Forward Selection, Support Vector Machines | Accuracy is 93.5 |
| 2010 | Geraudie V et al, | NIR Spectrometry | Averaging, Normalization and Transformation | - | Principal Component analysis, Multiple Linear Regression | - |
| 2010 | J.Liu et al | NIR Spectrometry | First Derivative, Vector Normalization, FD and VN, Second Derivative, SD and VN | - | Ward's Algorithm | Accuracy is 92.8 for slightly moldy to 100 |
| 2010 | Musa Atas et al, | Hyper spectral image | - | - | Principal Component analysis, Self associative neural network | Accuracy is 78 |
| 2011 | Hongfei Lu et al, | Multispectral | - | - | Principal Component analysis, support vector machine | Accuracy is 100 |
| 2011 | H.Kalkan et al, | Multispectral | Bandpass filters | - | Two Dimensional linear discriminant bases(LDB), local discriminant analysis(LDA), Principal Component analysis | Accuracy is 80 |

**Authors**

Dr. P. Subashini, 18 years of teaching experience-working as Associate Professor in Avinashilingam Deemed University for women, Area of Specialization: Image Processing, Pattern recognition, Neural Networks.
Email id : mail.p.subashini@gmail.com

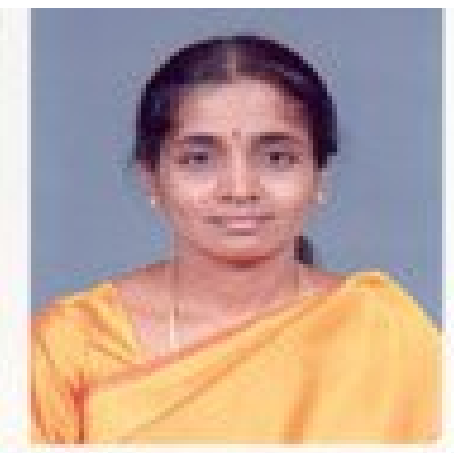

M.Rajalakshmi, 7 years and 6 months of teaching experience - working as a Guest Lecturer in Government Arts College, Salem - 7. Doing her research work in Image processing. Area of Specialization: Image Processing, Pattern recognition.

Email id: rajisaravanan25@gmail.com

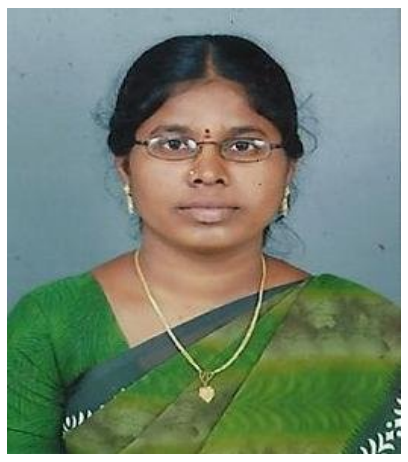